Ari Goodman, Ryan O'Shea
Naval Air Warfare Center Aircraft Division Lakehurst

# LIFT OFF: LoRaWAN Installation and Fiducial Tracking Operations for the Flightline of the Future

## Abstract

Real-time situational awareness for the location of assets is critical to ensure missions are completed efficiently and requirements are satisfied. In many commercial settings, the application of global positioning system (GPS) sensors is appropriate to achieve timely knowledge of the position of people and equipment. However, GPS sensors are not appropriate for all situations due to flight clearance and operations security concerns. LIFT OFF: LoRaWAN Installation and Fiducial Tracking Operations for the Flightline of the Future proposes a hybrid framework solution to achieve real-time situational awareness for people, support equipment, and aircraft positions regardless of the environment. This framework included a machine-vision component, which involved setting up cameras to detect AprilTag decals that were installed on the sides of aircraft. The framework included a geolocation sensor component, which involved installing GPS sensors on support equipment and helmets. The framework also included creating a long-range wide area network (LoRaWAN) to transfer data and developing a user interface to display the data. The framework was tested at Naval Air Station Oceana Flightline, the United States Naval Test Pilot School, and at Naval Air Warfare Center Aircraft Division Lakehurst. LIFT OFF successfully provided a real-time updating map of all tracked assets using GPS sensors for people and support equipment and with visual fiducials for aircraft. The trajectories of the assets were recorded for logistical analysis and playback. Future follow-on work is anticipated to apply the technology to other environments including carriers and amphibious assault ships in addition to the flightline.

## Introduction

Current tracking operations for flightline assets such as people, aircraft, and support equipment, are performed through manual, visual inspection. There is no central log for the position of these assets, nor is there a capability to assess their real-time position besides line-of-sight verification. Relying on visual asset tracking on Navy flight lines inhibits efficiency.

Readiness, sortie generation rate, and safety could be improved through enhanced situational awareness of command and control elements. Industry standard alternatives to visual asset tracking have not solved all these issues because placing tracking devices on equipment creates electronic emissions. They also can require configuration changes to hardware (i.e., technical directives) which limits their applicability to Navy aircraft. There exist new technology solutions to allow passive tracking of assets using computer vision software, as well as commercial-off-the-shelf hardware to track personnel and support equipment.

In this work, two methods are presented which provide situational awareness for tracking assets on the flightline. In the first method, hardware and software were developed to track the position of people and support equipment with GPS sensors over a LoRaWAN installation. In the second method, aircraft were tracked using passive computer vision software and sticker decals called AprilTags.

## Background

Global Positioning System (GPS) tracking systems are used by companies to track the location and movement of assets, such as delivery vehicles and packages. GPS tracking systems use a network of satellites to determine the precise location of a device on the ground. GPS tracking systems can be used for a variety of purposes, including fleet management, asset tracking, and logistics. These systems allow real-time visibility and awareness of the current position and expected arrival time of packages [1]. The widespread use of GPS and ethical implications of its use have been explored in [2].

AprilTags are a popular marker-based visual fiducial system, developed by the University of Michigan. AprilTags are small, distinctive black and white square tags that can be attached to objects or printed on flat surfaces. They are designed to be easily detected and recognized by machine vision algorithms, even when partially occluded or under challenging lighting conditions [3].

AprilTags are used for a variety of applications, including robotic localization and mapping, augmented reality, and object tracking. They are particularly useful for robotics applications because they can be easily detected and recognized at a distance, even when the robot is moving [3].

LoRaWAN (Long Range Wide Area Network) is a type of wireless communication technology that is designed for low-power, long-range communication. It is commonly used for Internet of Things (IoT) applications, such as asset tracking, smart city infrastructure, and remote sensing. LoRaWAN operates in the unlicensed radio frequency spectrum and uses a spread spectrum modulation technique called chirp spread spectrum (CSS) to transmit data over long distances with low power consumption. It is designed to provide bi-directional communication over a range of several kilometers, depending on the specific implementation and operating conditions.

One of the main advantages of LoRaWAN is its low power consumption, which allows it to be used with battery-powered devices that need to operate for long periods of time without the need for frequent battery replacements. It is also well suited for applications that require long-range communication, as it can transmit data over distances that are beyond the range of other wireless technologies such as WiFi or Bluetooth.

LoRaWAN is typically used in conjunction with gateways that connect the network to the Internet and allow devices to communicate with each other and with cloud-based applications. It is an open standard that is widely adopted for IoT applications [4].

## Methodology

The first phase of the effort involved developing and installing GPS tracking devices on a custom LoRaWAN network at NAS Oceana. The GPS positions would relay their position to users over a custom user interface on a laptop connected to the network as seen in Figure 1. The purpose of this phase was to maintain situational awareness for personnel and support equipment.

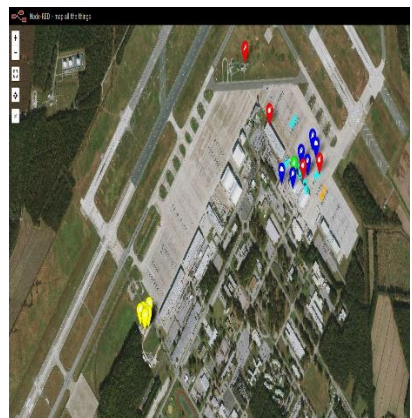

Figure 1: User Interface and Live Plot for GPS Locations

GPS trackers were purchased and installed in different locations depending on the application. For personnel, the tracker was installed in the helmet shown in Figure 2 that is commonly worn on the flight line. For support equipment like fuel trucks, a housing was 3D printed to store the electronics and attached to the wind shield as shown in Figure 3. In total, 22 GPS units were attached to various support equipment and personnel.

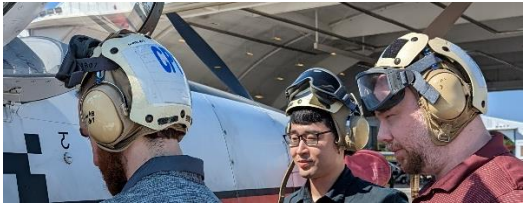

Figure 2: Helmets for Flightline Personnel

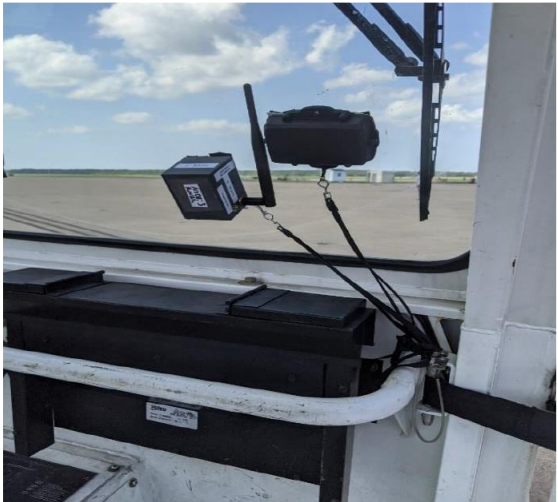

Figure 3: GPS Tracker on Windshield of Support Equipment

A LoRaWAN network was set up to collect data from the GPS trackers. A diagram of the network is shown below in Figure 4.

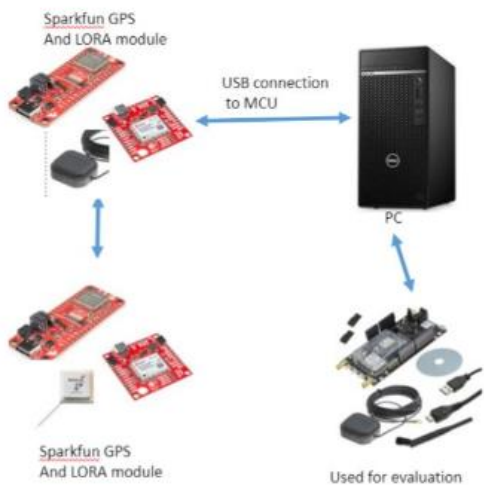

Figure 4: LoRaWAN Architecture

In the second phase, a prototype fiducial tracking system was created to track the real-time position of aircraft.

First, the environmental scope of the tracking system was established. These included the field of view, depth of field, and weather conditions. The field of view and depth of field were determined based on the testing environment at United States Naval Test Pilot School (USNTPS) depicted below in Figure 5.

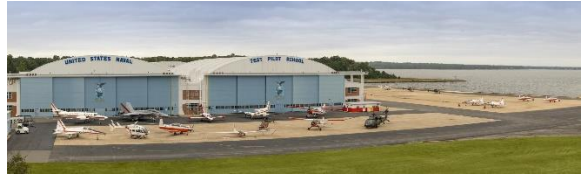

Figure 5: Overview of USNTPS

The weather conditions would exclude extreme weather, but could include variable sunny or cloudy conditions, or even light rain. During the week of testing, there was light rain and variable sunny and cloudy conditions.

The camera was selected to try to optimize for the maximum number of aircrafts captured in view. The Panasonic LUMIX DC-BGHI camera was used to capture video in this experiment. It is a box style camera that allowed for 4K recording of videos. The Rokinon 12mm T2.2 Cine Lens for Micro Four Thirds Mount lens was used in this experiment. This is a wide angled lens for digital cinematography. A wide angle lens was chosen to increase the field of view being covered by the camera.

The AprilTag family selected was 52h13 which was the most robust, freely available AprilTag family at the time.

AprilTag detection software was developed to track multiple AprilTags of various sizes. This software was integrated with a user interface to show the location of the aircraft relative to the camera. The information was stored for post processing, playback, and analysis. A direct linear transform (DLT) was used to calibrate the location of the camera system from a set of known points in the video.

The AprilTags were initially tested at Lakehurst on the Experimental Ground Vehicle (EGV) depicted below in Figure 6. Different materials

were used to generate different sized April Tags. Matte black and white tags were used for their high detection rate, and because these materials were approved to be used in experiments on aircraft at USNTPS.

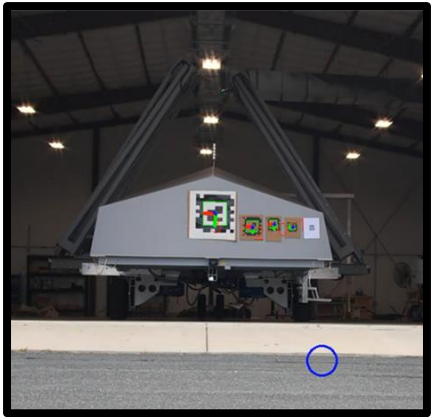

Figure 6: Experimental Ground Vehicle with multiple AprilTags and Detections

The experiments with the EGV validated the following equation used to estimate maximum detection distance:

$$Distance = \frac{t}{2 * tan(\frac{b * f * p}{2 * r})}$$

Max detection distance in meters [5]
t = size of tag in meters
b = number of bits that span the width of the tag
f = horizontal FOV
p = the number of pixels required to detect a bit
r = horizontal resolution

For these validating experiments, AprilTags were applied to the asset and the camera was moved away until the AprilTags were no longer detected.

The setup was further tested on real at USNTPS. Over the course of a week, multiple AprilTags of varying sizes were attached to multiple aircrafts as shown in Figure 7. An example tagged aircraft can be seen in Figure 8.

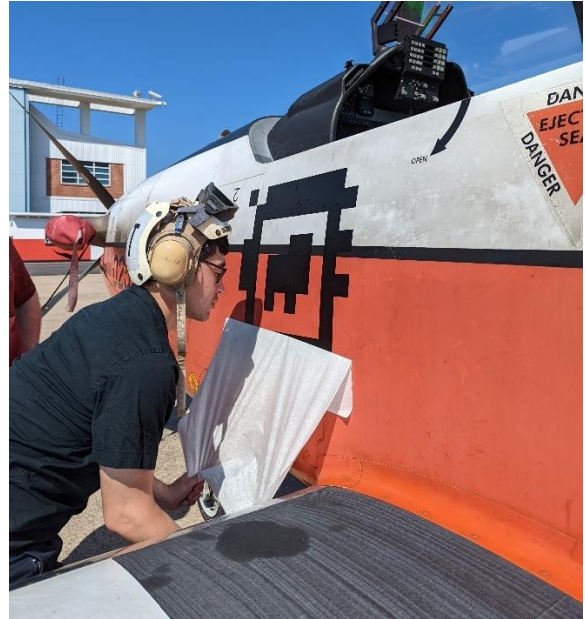

Figure 7: Applying AprilTag to Aircraft

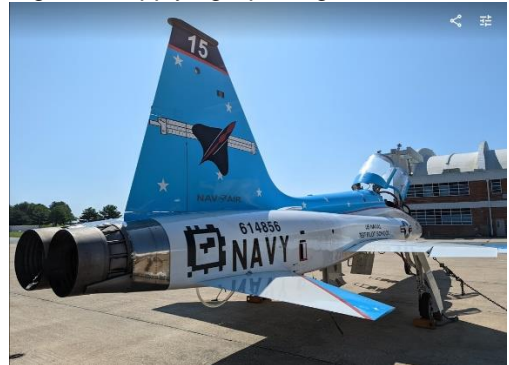

Figure 8: Aircraft with installed AprilTag

Next, the cameras were set up to detect the AprilTags while they taxied past the cameras as shown in Figure 9.

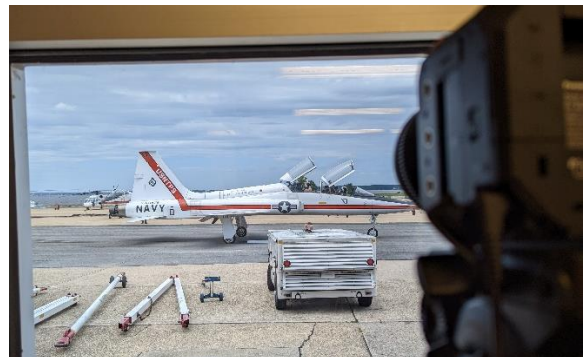

Figure 9: Camera Setup Capturing Aircraft Footage from Indoors

The footage was then processed, and detections were translated into 3D coordinate estimates without relying on other sensors such as GPS. The tags were removed with minimal impact on the aircrafts' exteriors at the end of the experiment.

## Results

The GPS sensors were able to report the real-time location of personnel and support equipment over the LoRaWAN network at NAS Oceana. The trajectories of all tagged assets were able to be replayed during post analysis as shown in Figure 10 below.

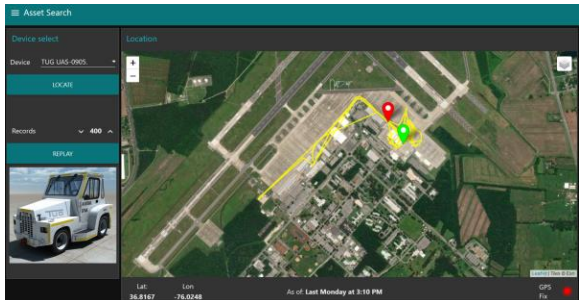

Figure 10: Example Replay of GPS Trajectory

Several decals were installed on multiple aircrafts at the United States Naval Test Pilot School. The application tools (squeegees) were used to ensure flat application. However, small air bubbles were still present upon close inspection. Tags of multiple sizes were used, depending on the size of the free space available on the aircraft. To meet flight criteria, the tags did not cover any existing control surfaces which limited their size and location.

Several videos were recorded at USNTPS and processed by the AprilTag detecting software. An example processed image is shown in Figure 11. The software was moderately successful at detecting the AprilTags, depending on the size of the tag, angle to the tag, and distance to aircraft. The light rain and glare from the sun did not have a significant impact on detection. In the future, it is recommended to have cameras with a stronger optical zoom to allow for detections at greater distances.

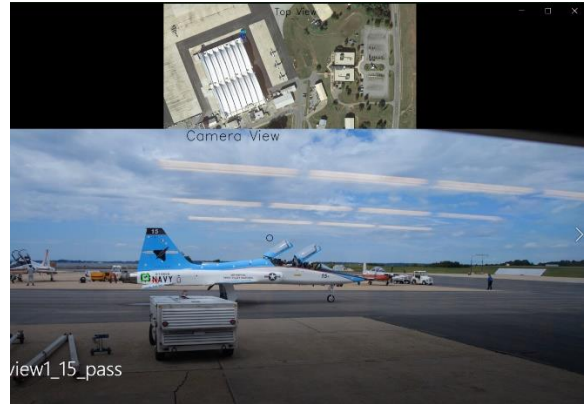

Figure 11: Aircraft Detection with Updating Map

Multiple AprilTag colors and materials were tested on the EGV. The color did not significantly impact performance. The application of the decal proved to be more critical than the color; a flat application with strong adherence was crucial to avoid warping. AprilTag detection occurred in real time at 1080p, but more computing power and parallelization is needed to process 4K videos in real time. The 4K videos were first recorded and processed afterwards.

The further the tags were from the camera, the more uncertain the tag location and pose estimates. The wide-angle lens allowed for capturing more aircraft in a single image, but the tag detections were less robust due to the increased field of view. A different lens with a greater focal length, or a second camera dedicated to zooming, would allow for more robust detection.

## Conclusion

Overall, this effort demonstrated the capability to track people and support equipment in real time using GPS sensors and a LoRaWAN network and aircraft in real time using AprilTags and COTS camera hardware.

COTS GPS sensors were successfully used to track personnel and support equipment over a custom LoRaWAN installation. The software showed the ability to have real-time awareness for the states of assets as well as provided the capability for virtual playback of past events.

Fiducial decals were successfully used to identify and locate aircraft. The methods used to manufacture, apply, track, and remove decals met requirements.

In the future, it is recommended to apply the dark AprilTags onto a white continuous background prior to applying the background and the tag onto to the aircraft together. This method would be easier and would remove colors or patterns pre-existing on the aircraft from interfering with the AprilTag Detection. It is recommended to use Pan Tilt Zoom (PTZ) cameras to have improve the maximum detection distance while maintaining thorough coverage of the flightline.

## Acknowledgments

The authors would like to acknowledge Alex Wendt and Christopher Thajudeen for their efforts in constructing and designing the LoRaWAN network and GPS sensors. The authors would also like to acknowledge Daniel Bramos for his effort in interfacing with NAS Oceana and USNTPS and managing the demonstrations. The authors would also like to acknowledge Jianyu An, Kevin Larkins, and Tushar Patel for helping perform experiments.

Ari Goodman is the S&T AI Lead and a Robotics Engineer in the Robotics and Intelligent Systems Engineering (RISE) lab at Naval Air Warfare Center Aircraft Division (NAWCAD) Lakehurst. In this role he leads efforts in Machine Learning, Computer Vision, and Verification & Validation of Autonomous Systems. He received his MS in Robotics Engineering from Worcester Polytechnic Institute in 2017.

Ryan O'Shea is a Computer Engineer in the Robotics and Intelligent Systems Engineering (RISE) lab at Naval Air Warfare Center Aircraft Division (NAWCAD) Lakehurst. His current work is focused on applying computer vision, machine learning, and robotics to various areas of the fleet to augment sailor capabilities and increase overall operational efficiency. He received a Bachelor's Degree in Computer Engineering from Stevens Institute of Technology.